\pdfoutput=1
\documentclass[11pt]{article}

\usepackage{acl}

\usepackage{times}
\usepackage{latexsym}
\usepackage{graphicx}
\usepackage{booktabs, multirow, makecell}
\usepackage{adjustbox}
\usepackage{dingbat}
\usepackage{pifont}
\usepackage{xcolor}
\usepackage{enumitem}
\usepackage[T1]{fontenc}
\usepackage[utf8]{inputenc}

\usepackage{microtype}

\usepackage{inconsolata}
\usepackage{comment}

\title{PRACTIQ: A Practical Conversational text-to-SQL dataset with Ambiguous and Unanswerable Queries}

\author{%
  \textbf{Mingwen Dong}$^\dagger{}^*$
  \ , \
  \textbf{Nischal Ashok Kumar}$^\ddagger$\thanks{Co-first authors with equal contribution. Work done while Nischal Ashok Kumar was an intern at AWS.}
  \\
  \textbf{Yiqun Hu$^\dagger$, Anuj Chauhan$^\dagger$, Chung-Wei Hang$^\dagger$, Shuaichen Chang$^\dagger$, Lin Pan$^\dagger$,}\\
  \textbf{Wuwei Lan$^\dagger$, Henghui Zhu$^\dagger$, Jiarong Jiang$^\dagger$, Patrick Ng$^\dagger$, Zhiguo Wang$^\dagger$}\\
  $^\dagger$Amazon Web Services, $^\ddagger$University of Massachusetts at Amherst
  \\
  \texttt{nashokkumar@umass.edu}, \texttt{\{mingwd, jiarongj, patricng, zhiguow\}@amazon.com}
}

\begin{document}
\maketitle
\begin{abstract}
Previous text-to-SQL datasets and systems have primarily focused on user questions with clear intentions that can be answered. However, real user questions can often be ambiguous with multiple interpretations or unanswerable due to a lack of relevant data. In this work, we construct a practical conversational text-to-SQL dataset called \emph{PRACTIQ}, consisting of ambiguous and unanswerable questions inspired by real-world user questions. We first identified four categories of ambiguous questions and four categories of unanswerable questions by studying existing text-to-SQL datasets. Then, we generate conversations with four turns: the initial user question, an assistant response seeking clarification, the user's clarification, and the assistant's clarified SQL response with the natural language explanation of the execution results. For some ambiguous queries, we also directly generate helpful SQL responses, that consider multiple aspects of ambiguity, instead of requesting user clarification. To benchmark the performance on ambiguous, unanswerable, and answerable questions, we implemented large language model (LLM)-based baselines using various LLMs. Our approach involves two steps: question category classification and clarification SQL prediction. Our experiments reveal that state-of-the-art systems struggle to handle ambiguous and unanswerable questions effectively. We will release our code for data generation and experiments on GitHub\footnote{https://github.com/amazon-science/conversational-ambiguous-unanswerable-text2sql/tree/main}.
\end{abstract}

\begin{figure*}         
    \centering
    \includegraphics[width=\textwidth]{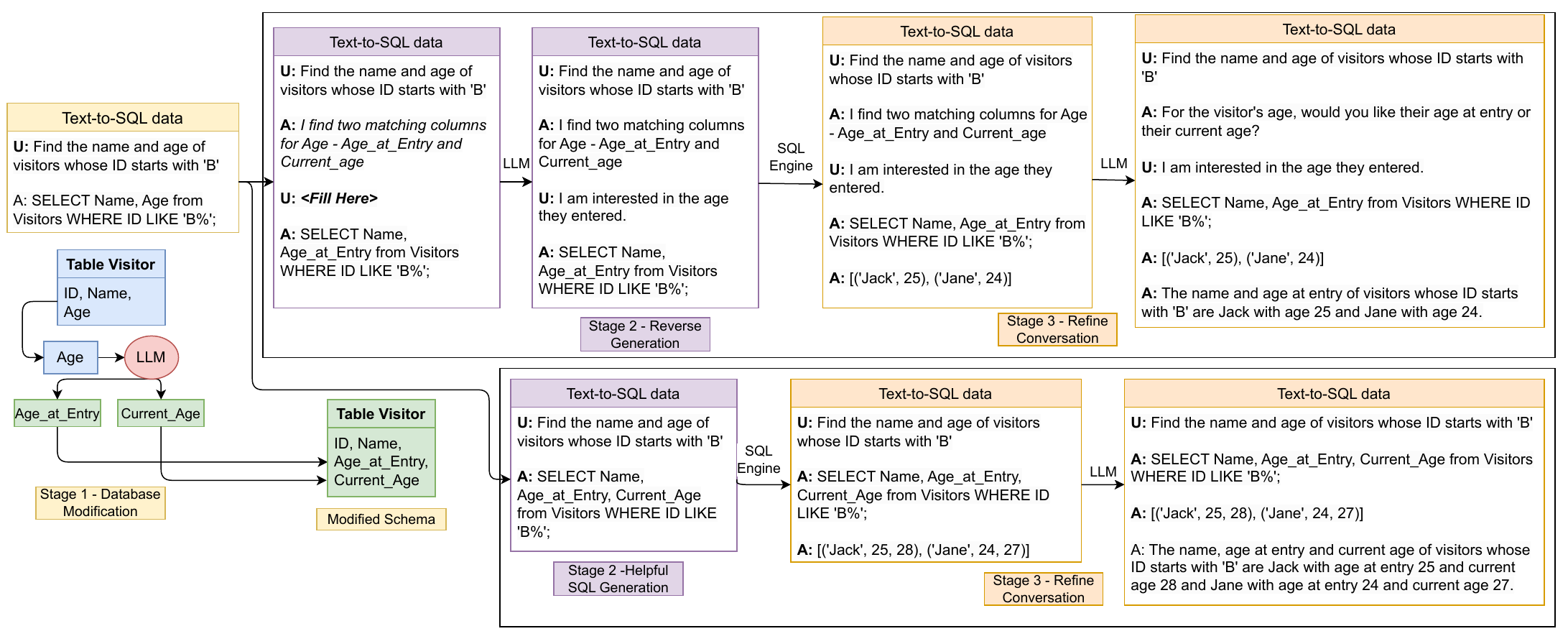}
    \caption{An example of our conversational dataset construction consists of three stages: database modification, SQL modification along with clarification response generation, and refining the conversation. The top box depicts our data construction for an ambiguous question that requires clarification questions, while the bottom box illustrates an ambiguous question with direct helpful SQL responses. Here `U' refers to a user and `A' refers to a text-to-SQL assistant.}
    \label{fig:intro-example}
\end{figure*}

\section{Introduction}

Text-to-SQL systems aim to convert natural language questions into SQL queries that can be used to query a database. The systems serve as an interface between users and databases to allow the users access to information from the databases through their natural language questions. The advent of Large Language Models (LLMs) \cite{bubeck2023sparks} has significantly enhanced the capabilities of text-to-SQL systems, such as DIN-SQL \cite{pourreza2024din}, achieving state-of-the-art (SoTA) performance on standard benchmarks\footnote{As of August 2023}, including Spider \cite{yu-etal-2018-Spider} and BIRD \cite{li2024can}.

Although the SoTA text-to-SQL systems perform well on clean benchmarks that contain only answerable user queries, they are still not well-equipped to deal with practical real-world data which have ambiguous or unanswerable questions \cite{wang2023know}. The poor performance of SoTA text-to-SQL systems is primarily due to the unavailability of practical text-to-SQL data that can be used for training \cite{wang2023know}. Although previous research finds that a large ratio of user questions are unanswerable, these are often excluded in the previous datasets as addressing them requires more than SQL annotations \cite{lee-etal-2021-kaggledbqa}.
To bridge this gap, we introduce \emph{PRACTIQ} which is a \textbf{practi}cal conversational text-to-SQL dataset with ambiguous and unanswerable \textbf{q}ueries. As illustrated in Table \ref{tab:ambi-unans-combined}, a question is ambiguous if it has multiple valid interpretations given the database schema and the question is unanswerable if the corresponding database does not contain the data that the question is asking for. In the real world, given a user question, a text-to-SQL assistant has to first determine whether the question is answerable, ambiguous, or unanswerable to decide whether to ask for clarification questions or respond with the correct SQL.

We begin by examining existing text-to-SQL datasets \cite{yu-etal-2018-Spider,li2024can,yu-etal-2019-cosql} and identify four ambiguous and four unanswerable categories inspired by real-world practical user questions. Subsequently, we generate ambiguous and unanswerable examples corresponding to these categories by parsing the SQLs and modifying the databases (Spider~\footnote{We only used ``train\_spider.json'' and ``dev.json'' from Spider in our dataset construction and related experiments. We did NOT use any data from ``train\_others.json'' or ``yelp''.} is used in the current work, but the framework can be easily adapted to other text-to-SQL datasets). We then leverage an LLM to convert the data into conversations between the user and a text-to-SQL assistant that includes user initial questions, assistant clarification questions, user clarification responses, assistant SQL responses, SQL execution results, and natural language explanations of the execution results (as shown in Figure \ref{fig:intro-example}). In addition to having conversations where the assistant asks for clarification questions, we also generated more helpful SQL responses that included the results of all possible responses for some ambiguous question categories. To assess the quality of our generated dataset we define annotation criteria for two tasks, question category classification, and conversation quality evaluation, and conduct human annotation on the generated data to show that our dataset is of high quality. Finally, we propose prompt-based baselines to benchmark our dataset on the text-to-SQL generation task, which involves two tasks, classifying the category of the user question, and then generating the clarification SQL based on the user question. We experiment with several SoTA LLMs and show that the current text-to-SQL systems still need improvements on real-world queries that include ambiguous or unanswerable questions.

Our contributions can be summarized as follows: 
\begin{itemize}[leftmargin=*]
    \setlength\itemsep{-0.4em}
    \item We study existing text-to-SQL datasets and identify four ambiguous and four unanswerable question categories inspired by real-world user questions. We implemented a framework and programmatically generated \emph{PRACTIQ}, a comprehensive and fine-grained ambiguous and unanswerable text-to-SQL dataset consisting of 2800 conversations.
    \item We extend the ambiguous/unanswerable data into conversations between a user and an assistant. The conversation typically includes a user initial question, a helpful assistant response seeking user clarification, a user clarification response, the assistant SQL response, SQL execution results, and a natural language explanation.
    \item To the best of our knowledge, our work is the first to study text-to-SQL systems when user queries can be answerable, ambiguous, or unanswerable in a conversational setting.
    We benchmark various SoTA LLMs on \emph{PRACTIQ} on two sub-tasks: 
    question category classification, and clarification SQL prediction. Our results show that the ambiguous and unanswerable questions are challenging even for methods leveraging SoTA LLMs indicating the need to improve LLMs' handling of real-world practical text-to-SQL data. 
\end{itemize}

\begin{table*}[htbp]
\centering
\caption{Table showing the comparison of our work with existing datasets on ambiguity and unanswerablility in text-to-SQL task. \checkmark signifies that the category is present in the dataset. \ding{55} signifies that the category is not present in the dataset. * signifies that the category is defined and data is generated, but ambiguities are defined from a different perspective.}
\label{tab:compare-related}
\tiny
\begin{tabular}{p{2.5cm}|p{1cm}p{1cm}p{1cm}p{1cm}p{1cm}p{1cm}p{1cm}p{1cm}|p{1cm}}
\toprule
 & Ambiguous SELECT Column  & Ambig. Values Within Column & Ambiguous WHERE Column & Ambiguous Filter Criteria & Nonexistent SELECT Column & Nonexistent WHERE Column & Nonexistent Filter Value & Unsupported Join & Conversational \\
\midrule
NoisySP~\cite{wang2023know} & \checkmark & \ding{55} & $\sim$ & $\sim$ & \checkmark & $\sim$  &  $\sim$ &  \ding{55} & \ding{55}\\
\hline
AmbiQT~\cite{bhaskar-etal-2023-benchmarking} & \checkmark & \ding{55} & \ding{55} & \ding{55} & \ding{55} & \ding{55} & \ding{55} & \ding{55} & \ding{55}\\
\hline
AMBROSIA~\cite{saparina2024ambrosia} & * & * & * & * & \ding{55} & \ding{55} & \ding{55} & \ding{55} & \ding{55}\\
\hline
\textbf{\emph{PRACTIQ}} (ours) & \checkmark & \checkmark & \checkmark & \checkmark & \checkmark & \checkmark & \checkmark & \checkmark & \checkmark\\
\bottomrule
\end{tabular}
\end{table*}

\begin{table*}[!htb]
\tiny
\centering
\caption{Definition and Example of Ambiguous and Unanswerable Categories. Note that the question and database schema are simplified for illustration purposes.}
\label{tab:ambi-unans-combined}
\begin{tabular}{{p{0.1\linewidth} p{0.15\linewidth} p{0.7\linewidth}}}
\toprule

Category & Definition & Example \\ \midrule

Ambiguous SELECT Column  & Question with multiple valid SQLs that differ in the columns used in the SELECT clause. &  \textbf{Database Schema}:\newline
\hspace*{0.3cm}Stadium: Stadium Name, \textcolor{blue}{Standing Capacity}, \textcolor{blue}{Seating Capacity}, Average\_Num\_Games\_Played \newline
\textbf{Question}: What is the maximum \textcolor{orange}{capacity} of all stadiums? \newline
\textbf{Ambiguity}: There are two Ambiguous SELECT Columns - \textcolor{blue}{standing capacity} and \textcolor{blue}{seating capacity}.
\\ \midrule

Ambiguous Values Within Column & Questions that can map to selecting rows that correspond to multiple different values in the table &  \textbf{Database Schema}:\newline
\hspace*{0.3cm}Classroom: Subject, Teacher Name, Number of Students Enrolled 
\newline
\textbf{Question}: Who is the \textcolor{orange}{Chemistry} teacher? 
\newline
\textbf{Ambiguity}: The table contains two possible chemistry values in the Subject column: \textcolor{blue}{Organic Chemistry} and \textcolor{blue}{Physical Chemistry}.
\\ \midrule

Ambiguous WHERE Columns & Questions that can map to selecting rows that correspond to the same value in multiple different columns &  \textbf{Database Schema}:\newline
\hspace*{0.3cm} Properties:  \textcolor{blue}{property\_type\_code}, \textcolor{blue}{property\_type\_version}, properties description, property\_name, room count; 
\newline \textbf{Question}: What are the names of properties whose \textcolor{orange}{property type} is a multiple of 5? 
\newline \textbf{Ambiguity}: Both \textcolor{blue}{property\_type\_code} \& \textcolor{blue}{property\_type\_version} column contain cell value 5.
\\ \midrule

Ambiguous Filter Criteria  & Questions containing terms that definition/mapping of values in the database &  \textbf{Database Schema}:\newline
\hspace*{0.3cm} Thrombosis\_Prediction: \textcolor{blue}{patient age}, date, patient\_id, examined\_or\_not
\newline \textbf{Question}: How many \textcolor{orange}{underage} patients were examined during the three years from 1990 to 1993?
\newline \textbf{Ambiguity}: Underage is ambiguous: it means younger than a certain age but what specific age can differ and require definition.
\\ \midrule 

Nonexistent\newline SELECT Column & The column that contains the results asked in the question do not exist in the database &  \textbf{Database Schema}:\newline
\hspace*{0.3cm} Olympics: Medal, Name of Sportsman, Sport, Event
\newline \textbf{Question}: What was the \textcolor{orange}{nickname} of the gold medal winner in the men's heavyweight greco-roman wrestling event of the 1932 Summer Olympics?
\newline \textbf{Unanswerability}: The table does not contain any information on nicknames.

\\ \midrule
Nonexistent WHERE Column  & Column(s) asked for filtering the information in the question do not exist in the database &  
\textbf{Database Schema}:\newline
\hspace*{0.3cm} Teams: Team Name, Ground, Town Name, Previous Standing
\newline \textbf{Question}: Which team of the Cornwall League 1 comes from a \textcolor{orange}{town} that is known for its \textcolor{orange}{tin mining}?
\newline \textbf{Unanswerability}: The table does not have any information about tin mining and there are no columns containing information that defines tin mining (different from Ambiguous Filter Criteria ambiguous).

\\ \midrule

Nonexistent Filter Value  & Questions that ask for value(s) not present in the database & \textbf{Database Schema}:\newline
\hspace*{0.3cm} Teams: Team Name, Ground, Town Name, Previous Standing
\newline \textbf{Question}: What is the ground name of \textcolor{orange}{New York Yankees}?
\newline \textbf{Unanswerability}: The table does not have any information about the New York Yankees in the team name column.
\\ \midrule

Unsupported Join  & Questions that ask information covering tables in the database that cannot be joined (are not connected by foreign keys) &  \textbf{Database Schema}:\newline
\hspace*{0.3cm} Tables - Students, Teachers, Grades,..., and Library, and Books; Here students, teachers, and grades columns are connected using foreign keys but not to library and books.
\newline \textbf{Question}: Which \textcolor{orange}{student }borrowed the \textcolor{orange}{book} titled ``ABC'' from the \textcolor{orange}{library} ``XYZ''?
\newline \textbf{Unanswerability}: To answer this question, we need to join the student table with library-books tables. This JOIN operation is not possible as there are no overlapping columns or foreign keys that connect the two tables. 

\\ \bottomrule
\end{tabular}
\end{table*}

\section{Related Work}

\subsection{Standard text-to-SQL datasets}

Most text-to-SQL datasets, such as Spider \cite{yu-etal-2018-Spider}, BIRD \cite{wang2023know}, and WikiSQL \cite{zhong2017seq2sql}, consist of non-conversational, answerable questions with clear intent. SPARC and CoSQL are conversational but only have a very limited number of ambiguous or unanswerable questions \cite{finegan-dollak-etal-2018-improving, Yu&al.19, yu-etal-2019-cosql}. E.g., CoSQL contains around 10k annotated SQL queries from 3k dialogues spread across 200 complex databases, but there are only approximately 190 unanswerable questions and only 34 (approximately 18\%) of them request a user clarification to resolve the issue in the next turn. Also, the responses by the text-to-SQL system to such questions are not always helpful. For example, responses like ``Sorry, I can't answer this question using SQL.'' do not specify the exact reason why the question cannot be answered, which can discourage the users from asking follow-up questions. The ambiguous/unanswerable questions in CoSQL are not categorized into fine categories, probably due to the small size of such questions (12\% of the whole dataset). Our work fills this gap by generating a large number of ambiguous/unanswerable questions using eight different methods. With the advent of LLMs, there has been a wider focus on conversational dialogue-oriented systems that can engage with users helpfully to solve a particular task (\cite{wang-etal-2023-umass}, \cite{zhang-etal-2023-sgp}, \cite{deng-etal-2023-prompting}). We convert our data into conversational forms leveraging reverse generation (generating SQL first and then generating user clarification responses) using an LLM (see Figure \ref{fig:intro-example}).

\subsection{Ambiguity and Unanswerability in text-to-SQL systems}
Recent research has identified the presence of ambiguous and unanswerable questions in practical text-to-SQL systems. However, they primarily focused on creating ambiguous or unanswerable data to train question classifiers \cite{zhang2020did} or covered only a limited range of ambiguous/unanswerable categories \cite{wang2023know}. Concurrently, \citet{bhaskar-etal-2023-benchmarking} introduced AmbiQT, a benchmark containing ambiguous text-to-SQL queries spanning four ambiguous categories, and suggested generating multiple SQL queries to encompass the correct SQL. More recently, AMBROSIA defined and generated ambiguous text-to-SQL data based on scope ambiguity, attachment ambiguity, and vagueness but did not cover unanswerable categories \cite{saparina2024ambrosia}. Text2Analysis \cite{he2024text2analysis} focuses on structured data and also includes unclear queries, however, its queries focus more on advanced analysis skills rather than text-to-SQL. Our dataset differs in several key aspects. First, we address more comprehensive and fine-grained categories considering both ambiguous and unanswerable queries. Second, \emph{PRACTIQ} extends the generated data into a conversational format, reflecting the resolution of the problem in the original user query through interactions, resembling practical settings. Lastly, we handle cases with ambiguous inputs that can be addressed without explicitly needing a user clarification by directly generating helpful SQL and natural language responses covering all ambiguous columns in the database for the Ambiguous SELECT Column and Ambiguous WHERE Column questions. Table \ref{tab:compare-related} compares the ambiguous and unanswerable categories defined in our work with existing datasets, highlighting the range of broader categories covered in \emph{PRACTIQ}. By addressing the limitations of existing datasets and providing a comprehensive and conversational dataset, our work aims to support the development of practical text-to-SQL applications that can handle ambiguous and unanswerable queries more effectively.

\section{Question Categorization \& Dataset Construction}

We analyzed public text-to-SQL datasets like Spider \cite{yu-etal-2018-Spider}, BIRD \cite{li2024can}, CoSQL \cite{yu-etal-2019-cosql} and proposed four ambiguous and four unanswerable categories, as shown in Table \ref{tab:ambi-unans-combined}. The ambiguous categories include Ambiguous SELECT Column, Ambiguous WHERE Columns, Ambiguous Values Within Columns, and Ambiguous Filter Criteria. Ambiguous questions have multiple possible interpretations and subsequently multiple correct SQL responses given the database schema. The unanswerable categories include Nonexistent SELECT Column, Nonexistent WHERE Column, Nonexistent Filter Value, and Unsupported Join. Unanswerable questions are those for which a valid SQL cannot be produced given the database schema.

The data generation process consists of three main stages, as shown in Figure \ref{fig:intro-example}. We describe the main procedure and illustrate it with a detailed explanation for one category. For convenience, we use "assistant" to indicate the text-to-SQL system in the remaining text. Please see Appendix \ref{appendix-sec:dataset-construction} for a detailed explanation of the data generation process for each category.

\subsection{Stage 1: SQL parsing \& Database modification}
We first extract the columns and cell values by parsing the SQL queries using a custom parser on top of SQLGLOT\footnote{https://github.com/tobymao/sqlglot}. Then, we select a column or cell value of interest and modify the database schemas using an LLM so that the question becomes ambiguous or unanswerable. Since users are often unaware of database details, modifying the databases instead of the user questions, when plausible, is a natural way to create ambiguous and unanswerable questions.
For example, for Ambiguous SELECT Column questions, we asked the LLM to generate two alternative columns to replace the original column mentioned in the question, such that either column is a valid interpretation of the question (see Prompt \ref{prompt:generating_replacement_select_columns} for details). For Nonexistent Filter Value questions, we remove the mentioned cell values from the database, making the question unanswerable. For example, given the user question "What is the maximum capacity of all stadiums?" and the original database schema with the column "Capacity", we prompt the LLM to generate two semantically similar but non-equivalent columns, "Standing Capacity" and "Seating Capacity". We then remove the original "Capacity" column and add the newly generated columns to the database.

\subsection{Stage 2: SQL modification and clarification response generation}
Based on the user question, the modified database, and the original SQL, we generate the text-to-SQL assistant's initial response to the ambiguous/unanswerable question, the following user clarification response, and the assistant's SQL response to the clarified question. First, we generate the assistant's response to the initial user question using either a template-based method or a prompting method. For example, for Ambiguous SELECT Column questions, the template is "I find two conflicting information in the data. Which one would you like to know about? \textit{Ambiguous\_SELECT\_Column\_1} or \textit{Ambiguous\_SELECT\_Column\_2}".

Next, we follow a reverse-generation process \cite{hu-etal-2023-importance} to first generate the assistant's final SQL response and then generate the user's clarification question. The assistant's final SQL response is generated by modifying the original SQL programmatically. Then, we prompt the LLM to fill in the user's clarification response based on the conversation context (initial user question, assistant's clarification question, and final SQL responses). For example, for the Ambiguous SELECT Column question, we generate the assistant's clarified SQL by replacing the column in the SELECT clause of the original SQL with one of the ambiguous SELECT columns generated in the above stage. Then, given the user's initial question, the assistant's clarification question, "\textit{empty\_user\_clarification\_response}", and the assistant's final SQL response, we prompt the LLM to fill in the "\textit{empty\_user\_clarification\_response}" so that the user clarification response matches the assistant's SQL response and rest of the conversation (see Prompt \ref{prompt:generating_user_clarification_response} for details). This process ensures that the assistant's clarified SQL is more accurate and executable, as we are not prompting the LLM to generate it, which could lead to incorrect SQL. Finally, we execute the constructed clarification SQLs against the modified databases and discard examples that are not executable. After the reverse generation and filtering, each sample includes the user's initial question, the assistant's clarification question, the user's clarification response, the assistant's SQL response, and its corresponding execution results.

\subsubsection{Generating helpful SQL for ambiguous questions}
Because it is not always helpful for the assistant to ask clarification questions for ambiguous/unanswerable queries, we also generate helpful SQL responses to the Ambiguous SELECT Column and Ambiguous WHERE Column queries and reversely generate the corresponding assistant's explanation of why the SQL response is helpful. For Ambiguous SELECT Column queries, we sometimes can simply return all valid interpretations of the columns in the SQL. For example, suppose the question "What is the maximum capacity of all stadiums?" is ambiguous because capacity can map to either "Standing Capacity" or "Seating Capacity". In that case, we can return both capacity columns, reducing the number of turns for the user to get the information they need. We only generate such helpful SQL responses for the Ambiguous SELECT Column and Ambiguous WHERE Column categories, but this can be extended to other categories in the future.

\subsection{Stage 3: Refining the conversation \& Quality Control}
Leveraging an LLM, as a post-processing step \cite{wang-etal-2023-umass}, we use a 3-shot prompt to improve the naturalness and coherence of the conversation and add a natural language explanation of the final SQL execution results (see Prompt \ref{prompt:refine_conversation_prompt} \& \ref{prompt:generate_execution_results_explanation} for details). We randomly select 3 examples of the original conversation (as obtained from Stage 2), rewrite it more naturally and coherently, and add a natural language explanation of the execution results.

In addition to the main steps for generating the data, we employ a separate evaluation step after each generation step to control the data quality besides optimizing the generation prompt. The filtering step uses both LLM and execution checks. The LLM is often used to evaluate the quality of the data generated from the previous step or rank different candidates if multiple candidates have been generated. For example, for an ambiguous SELECT column question, suppose we have generated "Standing Capacity" or "Seating Capacity" as alternative columns for the question "What is the maximum capacity of all stadiums?". We will have a separate prompt and a few-shot examples for the LLM to evaluate whether these two columns are good candidates and make the question ambiguous. For execution checks, whenever we make a database change or generate modified SQLs, we execute these SQLs against the modified database to ensure the SQLs are executable.

Lastly, after generating data for each category, we prompted a LLM to perform binary classification on whether the provided question and modified database pair belonged to the designed category or not. This classification was based on the definition of the category and several human-curated examples (see Prompt \ref{prompt:binary_classification} for details). We only retained the examples that passed this binary classification, ensuring that the generated data accurately represented the intended ambiguous or unanswerable category.

\subsection{Dataset Statistics}

Table \ref{tab:data_stats_and_human_accuracy} shows the statistics of the dataset generated using the Spider dev set with Claude 3 sonnet. Note that the employed methodology can be seamlessly adapted to other text-to-SQL datasets like BIRD, WikiSQL, or any other synthetically generated answerable text-to-SQL corpora combined with any LLM (e.g., Llama3.1 or mixtral). The generated dataset consists of 1,802 ambiguous and unanswerable questions spanning various categories. Additionally, we collected 1,034 answerable queries from the Spider dev dataset and augmented them with natural language explanations derived from their execution results. Consequently, our dataset encompasses 2,812 conversations in total.

\subsection{Human Annotation}
We performed human annotations on two tasks: question category classification and overall conversation quality evaluation (see Appendix~\ref{appendix_human_annotation_procedure} for details). Four SQL experts with at least a bachelor's degree in Computer Science or equivalent work experience in the United States served as annotators.

For the question category classification task, we sampled 20 question-database pairs for each category. Annotators classified these pairs in two ways:

\textbf{1. Binary classification}: Annotators classified whether the pair belonged to the respective category based on the definition (Table~\ref{tab:ambi-unans-combined}).

\textbf{2. 9-way classification}: Annotators classified the pair into one of the nine categories based on the definition (Table~\ref{tab:ambi-unans-combined}).

Table~\ref{tab:data_stats_and_human_accuracy} shows that the average binary classification accuracy was 93.75\%. Figure~\ref{fig:question_category_classification} indicates that the average 9-way classification accuracy was 88.33\% (see Figure~\ref{fig:cm_human_evaluation} for more details). These human annotation results suggest that our dataset is of good quality.

For the conversation quality evaluation, we define three criteria: 

\textbf{factuality}: measures how well the SQL query provided by the assistant correctly answers the user question;

\textbf{helpfulness}: measures how helpful the assistant's responses are in fully understanding the user's intent;

\textbf{naturalness}: rates how natural and fluent the conversation is. 

We sample 10 conversations per category to include 90 conversations. Each conversation is annotated by 2 different SQL experts with the same qualifications as mentioned above. The annotators rate each category on a Likert scale between 1 and 5, where 1 denotes perfect quality and 5 denotes the worst quality for every criterion.  

The human annotation results (Table~\ref{tab:summary_annotation_scores}) show that our dataset is of high quality, with good naturalness, helpfulness, and factuality score (see Appendix \ref{ap:human_conv_annotation} for more details).

\begin{table}[htbp]
\centering
\caption{Dataset statistics and human annotation accuracy on 20 samples per question type. "\#Ex" column shows the number of examples generated for each category. "Acc" column shows average binary classification accuracy from human expert.}
\label{tab:data_stats_and_human_accuracy}
\small
\begin{tabular}{lrr}
\toprule
Category & \#Ex & Acc \\
\midrule
Ambiguous SELECT Column & 171 & 90\% \\
Ambiguous WHERE Column & 105 & 90\% \\
Ambiguous Filter Criteria & 303 & 100\% \\
Ambiguous Values Within Column & 122 & 80\% \\
\midrule
Nonexistent SELECT Column & 482 & 95\% \\
Nonexistent WHERE Column & 236 & 95\% \\
Unsupported Join & 213 & 100\% \\
Nonexistent Filter Value & 170 & 100\% \\
\midrule
Answerable (Spider Dev Set) & 1034 & 100\% \\
\midrule
Total & 2812 & - \\
Avg (excl. answerable) & - & 93.75\% \\
\bottomrule
\end{tabular}
\end{table}

\begin{table}[!htb]
\centering
\small
\caption{Summary of Human Annotation Scores for Naturalness, Factuality, and Helpfulness.}
\label{tab:summary_annotation_scores}
\begin{tabular}{l|c c c}
\hline
\textbf{Category} & \textbf{Mean} & \textbf{Std} & \textbf{Krippendorff's Alpha} \\ \hline
Naturalness & 1.57 & 0.87 & 0.8207 \\ \hline
Factuality & 1.15 & 0.53 & 0.6829 \\ \hline
Helpfulness & 1.41 & 0.74 & 0.7602 \\ \hline
\end{tabular}
\end{table}

\section{Evaluation Task and Baselines}

In this section, we describe the two evaluation tasks and corresponding metrics. 
\begin{enumerate}
    \item Question category classification: classify whether the question is answerable or one of the 8 ambiguous/unanswerable categories (9-way classification). We use classification accuracy for the ambiguous and unanswerable categories to measure the classification performance.  
    \item Clarification SQL Generation: predict the final SQL given the assistant's clarification question and user's clarification response. We use execution accuracy to measure the model performance \cite{li2024can}.
\end{enumerate}

\subsection{Question Category Classification}
We employ a few-shot prompting strategy for the question category classification task, experimenting with various numbers of shots (0-3) and different LLMs via the litellm\footnote{https://github.com/BerriAI/litellm} library as a baseline method. The prompt contains the definition of every category along with a variable number of in-context examples per category (see Prompt \ref{prompt:nine_way_classification} \& \ref{prompt:example_user_assistant_message} for details). Each example includes an input comprising the initial user question and relevant cell values retrieved via a fuzzy matching approach, as described in \cite{lin-etal-2020-bridging, wang-etal-2020-rat} (denoted by ``lexicalOnly''). The in-context demonstrations contain human-curated step-by-step thoughts and classification of the question categories \cite{wei2022chain}. To evaluate the impact of cell value retrieval on classification accuracy, we include a setting where oracle (perfect) cell values are provided to the model (denoted by ``lexicalAndOracle''). This setting allows us to assess how well the model performs if cell value retrieval is perfect.

\begin{figure}
    \centering
    \includegraphics[width=0.49\textwidth]{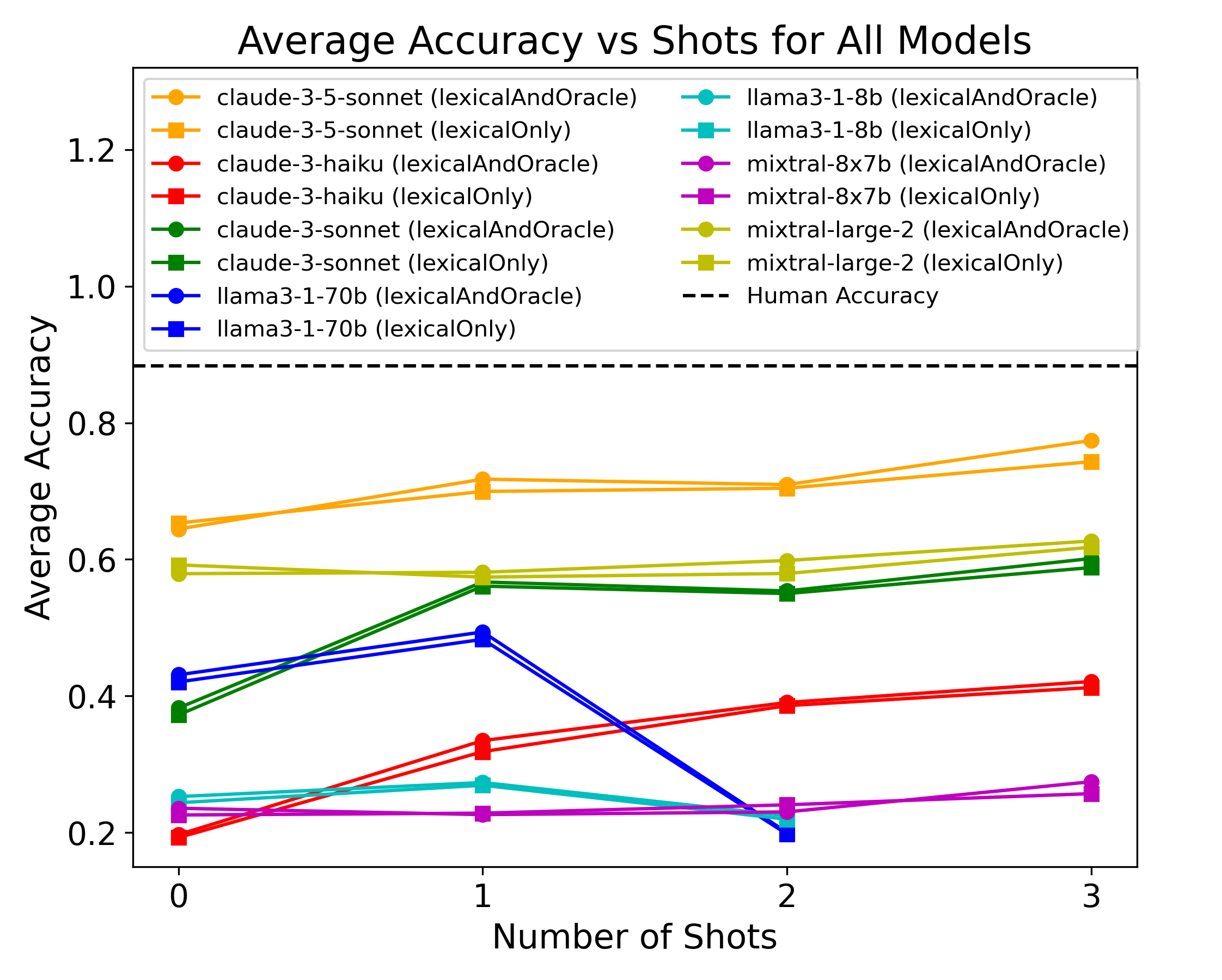}
    \caption{Figure showing the classification accuracy of different models using different number of shots.}
    \label{fig:question_category_classification}
\end{figure}

\subsection{SQL Prediction}
We use the DIN-SQL prompt-based framework, a SoTA method on the Spider dataset for predicting the final clarification SQL \cite{pourreza2024din}. The framework takes as input user questions and the corresponding database schema and contains four modules that decompose the task of SQL generation into several sub-tasks following a chain-of-thought \cite{wei2022chain} approach for SQL generation.

\begin{table*}[htbp]
\tiny
\centering
\caption{Execution accuracy of SQLs predicted with DIN-SQL using different LLMs on various categories of ambiguous, unanswerable, and answerable questions. The "All" column shows the overall average accuracy across all categories, while the "Avg. Excluding Answerable" column shows the average accuracy excluding the answerable questions from the Spider dataset.}
\label{tab:sql_accuracy_results}
\begin{tabular}{p{1.52cm}|p{0.7cm}p{0.8cm}p{1.0cm}p{0.7cm}p{0.7cm}p{0.8cm}p{0.7cm}p{0.8cm}p{0.86cm}|p{0.6cm}|p{1.4cm}}
\toprule
Model & Ambig. Filter Criteria & Ambig. SELECT Column & Ambig. Values Within Column & Ambig. WHERE Column & Nonexist. Filter Value & Nonexist. SELECT Column & Nonexist. WHERE Column & Unsupported Join & Answerable & Average & Avg. Excluding Answerable\\
\midrule
Claude 3.5 Sonnet & \textbf{77.23\%} & 67.25\% & 68.03\% & \textbf{77.14\%} & \textbf{74.12\%} & 64.73\% & 65.11\% & 76.53\% & \textbf{79.21\%} & 72.15\% & 71.27\%\\
Claude 3 Sonnet & 61.72\% & 58.48\% & 53.28\% & 59.05\% & 64.71\% & 55.19\% & 51.06\% & \textbf{77.46\%} & 64.12\% & 60.56\% & 60.12\%\\
Llama-3.1 70B & 68.65\% & 71.35\% & 63.11\% & 71.43\% & 65.88\% & \textbf{67.01\%} & \textbf{69.36\%} & 63.85\% & 76.31\% & 68.55\% & 67.58\%\\
Llama-3.1 8B & 48.84\% & 55.56\% & 45.90\% & 59.05\% & 54.71\% & 48.76\% & 46.81\% & 56.34\% & 56.58\% & 52.50\% & 52.00\%\\
Mixtral-large-v2 & 75.91\% & \textbf{74.27\%} & \textbf{69.67\%} & 75.24\% & 71.76\% & 66.18\% & 65.53\% & 77.00\% & 78.72\% & \textbf{72.70\%} & \textbf{71.95\%} \\
\bottomrule
\end{tabular}
\end{table*}

\section{Results and Discussions}

Figure \ref{fig:question_category_classification} shows the question category classification accuracy of different LLMs using varying numbers of examples. Claude 3.5 Sonnet\footnote{\url{https://www.anthropic.com/news/claude-3-5-sonnet}} achieves the best accuracy of 77.4\% (75.9\% excluding answerable category) across all categories when Oracle cell values are included in the schema and 3 examples per question type are provided. Without oracle cell values, the accuracy drops to 74.3\% (72.4\% excluding answerable). Mixtral-large-v2\footnote{\url{https://mistral.ai/news/mistral-large-2407/}} performs similarly to Claude 3 Sonnet when at least 1 example is provided per category but outperforms other models in the zero-shot setting, except Claude 3.5 Sonnet. For the average accuracy across all categories, having lexical cell values improves performance by 0.7\%, although the results are mixed. Across the three subcategories where cell values play a significant role (ambiguous VALUES within column, ambiguous WHERE column, and ambiguous filter criteria), having oracle cell values boosts classification accuracy by 1.5\%. These results show that improving cell value retrieval can be an important thing for detecting ambiguous/unanswerable questions in a practical text-to-SQL system, which previous research has not focused much on. 

The open-source Llama-3.1 70B \cite{touvron2023llama} model performs better than Mixtral-8x7b \cite{jiang2024mixtral} and Claude 3 Haiku but exhibits repeated text output when 2 or more examples are provided, causing its performance to drop below 20\%\footnote{With 3 examples per category, the results are similar, and the evaluation was stopped early for Llama-3.1 70B.}. These results indicate that detecting fine-grained ambiguity/unanswerability in questions given a database remains challenging for most LLMs (accuracy < 60\%), except for the powerful model Claude 3.5 Sonnet.

Table \ref{tab:sql_accuracy_results} shows our baseline method's (DIN-SQL) performance on SQL prediction of various LLMs given the interaction between the user and the assistant. Overall, Mixtral-large-v2 and Claude 3.5 Sonnet achieve the highest average accuracy of 71.95\% and 72.15\% on the ambiguous/unanswerable questions. Claude 3.5 sonnet achieves the highest performance of 79.21\% on the answerable questions (original Spider dev set). The open-source model Llama-3.1 70B performs competitively on the answerable questions achieving 76.31\% accuracy, only 2.9\% lower than Claude 3.5 sonnet. However, it performs only at 67.58\% accuracy on ambiguous/unanswerable questions, lagging 3.7\% behind Claude 3.5 sonnet. The gap can be as large as 9\% for some specific ambiguous question categories, indicating room for improvement. Our framework can be used to generate training data to improve open-source models' capabilities in both SQL prediction and detecting ambiguous/unanswerable questions.

\section{Conclusion and Future work}
In this work, we study current public text-to-SQL datasets and define four ambiguous and four unanswerable categories. We propose a framework to construct a practical conversational text-to-SQL dataset, \emph{PRACTIQ}, using both carefully constructed rules and Large Language Models (LLMs). We use the Spider dev dataset for constructing \emph{PRACTIQ} and generate around 2,800 conversational data samples. We evaluate our dataset on two core tasks, question category classification, and SQL prediction, and benchmark it using several SoTA LLMs.

Our results show that although some SoTA LLMs are approaching human-level accuracy, they are far from being perfect. For open-source models, the gap from human performance is much larger, indicating areas for further improvement. Our proposed framework provides a technique for generating additional practical text-to-SQL data on existing text-to-SQL datasets like WikiSQL, Spider Train, BIRD, or any other general synthetic single-turn answerable text-to-SQL data.
This practical enhancement of the datasets can be used to further train open-source models to enhance their capabilities in handling practical text-to-SQL tasks \cite{liu2024best}.

In a broader sense, our work presents a simple agentic workflow to generate synthetic data, which can be further used to improve LLMs. In the future, we can fine-tune open-source models with data generated using our framework to improve their capabilities. We can also experiment with agentic workflows to benchmark our dataset, and determine whether a question is ambiguous, unanswerable, or answerable, and accordingly provide more accurate and helpful responses.

\section*{Limitations}
While our dataset was generated using programmatic methods and LLMs, the data quality can be further improved by employing agentic workflows. Due to time constraints, we were unable to generate additional data to fine-tune open-source LLMs and evaluate whether fine-tuning can improve their ability to detect ambiguous/unanswerable questions and perform other reasoning tasks. We leave the exploration of fine-tuning open-source LLMs and the potential improvements in their capabilities as future work. We also encourage the research community to contribute to this effort by generating additional data using our proposed framework or exploring alternative approaches to enhance the quality and diversity of the dataset.

\section*{Ethics Statement}
Since we prompt LLMs on a large scale through a rate-based API for both dataset creation as well as evaluation the project may not be very environment friendly and may inevitably cause the emission of more CO2.

% Entries for the entire Anthology, followed by custom entries
\bibliography{custom,anthology}

\clearpage

\appendix

\section{Dataset Examples}
\label{sec:appendix}

\begin{table*}[ht]
\tiny
\centering
\caption{Table showing examples of the ambiguous category. Here U denotes the user and A denotes the assistant.}
\label{tab:ambi_example}
\begin{tabular}{p{3cm}|p{1cm}|p{0.5cm}p{10cm}}
\toprule
Ambiguous SELECT Column & Schema & & Table visitor, columns = [*,ID,Name,Level\_of\_membership,Age\_at\_Entry,Current\_Age]  \newline Table visit, columns = [*,Museum\_ID,visitor\_ID,Num\_of\_Ticket,Total\_spent] \newline Foreign\_keys = [visit.visitor\_ID = visitor.ID] \\ 
\cline{2-4}
& Conversation & U: & Find the name and age of the visitor who bought the most tickets at once. \\
& & A: & For the visitor's age, would you like the Age at Entry or Current Age? \\
& & U: & I'm looking for the age when they entered, so the Age at Entry. \\
& & A: & SELECT t1.name , t1.Age\_at\_Entry FROM visitor AS t1 JOIN visit AS t2  ON t1.id = t2.visitor\_id ORDER BY t2.num\_of\_ticket DESC LIMIT 1 \\ \midrule
Ambiguous Values Within Column & Schema & & Table Templates, columns = [*,Template\_ID,Version\_Number,Template\_Type\_Code, Date\_Effective\_From, Date\_Effective\_To, Template\_Details] \\
\cline{2-4}
& Relevant Values & & "templates.Template\_Type\_Code": ["useful CV 2", "useful CV 1", 'CV", "useful professional CV"] \\
\cline{2-4}
& Conversation & U: & How many templates have template type code useful CV? \\

& & A: & The value 'useful CV' matches multiple template type codes including 'useful CV 1', 'useful CV 2', and 'useful professional CV'.  Could you please specify which one you need the count for? \\

& & U: & How many templates have the template type code 'useful CV 2'? \\

& & A: & SELECT count(*) FROM Templates WHERE template\_type\_code = 'useful CV 2' \\
\midrule
Ambiguous WHERE Column & Schema & & Table battle, columns = [*,id,name,date,bulgarian\_commander,latin\_commander,result] \newline Table ship, columns = [*,lost\_in\_battle,id,name,tonnage,ship\_type,disposition\_of\_ship,Port\_of\_Origin,Destination] \newline Foreign\_keys = [ship.lost\_in\_battle = battle.id] \\
\cline{2-4}
& Relevant Values & & "ship.Port\_of\_Origin": ["English Channel"]; "ship.Destination": ["English Channel"] \\
\cline{2-4}
& Conversation & U: & Show names, results and bulgarian commanders of the battles with no ships lost in the 'English Channel'. \\

& & A: & I see 'English Channel' in two columns: Port of Origin and Destination. Can you clarify which you need? \\

& & U: & I meant the Port of Origin. \\

& & A: & SELECT name , result , bulgarian\_commander FROM battle EXCEPT SELECT T1.name , T1.result , T1.bulgarian\_commander  FROM battle AS T1 JOIN ship AS T2 ON T1.id = T2.lost\_in\_battle WHERE T2.Port\_of\_Origin = 'English Channel' \\
\midrule
Ambiguous Filter Criteria & Schema & & Table bond, columns = [*,bond\_id,molecule\_id,bond\_type] \newline Table molecule, columns = [*,molecule\_id,label] \newline Foreign\_keys =[bond.molecule\_id = molecule.molecule\_id] \\
\cline{2-4}
& Conversation & U: & What proportion of single bonds are carcinogenic? \\

& & A: & Could you clarify the meanings of 'single bond' and 'carcinogenic' in the context of your question? \\

& & U: & 'single bond' refers to bond\_type = '-'. 'carcinogenic' means label = '+'.  To calculate proportion: DIVIDE(SUM(label = '+'), COUNT(bond\_id)) where bond\_type = '-'. \\

& & A: & SELECT CAST(COUNT(CASE WHEN T2.label = '+' THEN T1.bond\_id ELSE NULL END) AS REAL) * 100 / COUNT(T1.bond\_id)  FROM bond AS T1 INNER JOIN molecule AS T2 ON T1.molecule\_id = T2.molecule\_id WHERE T1.bond\_type = '-' \\
\bottomrule
\end{tabular}
\end{table*}

\begin{table*}[ht]
\tiny
\centering
\caption{Table showing examples of the unanswerable category. Here U denotes the user and A denotes the assistant.}
\label{tab:unans_examples}
\begin{tabular}{p{3cm}|p{1cm}|p{0.5cm}p{10cm}}
\toprule
Nonexistent SELECT Column & Schema & & Table airports, columns = [*,City,AirportCode,Country,CountryAbbrev] \\
\cline{2-4}
& Conversation & U: & What is the airport name for airport 'AKO'? \\

& & A: & The airport name is not available in the schema. \\

& & U: & Can you provide the country for the airport code 'AKO' instead? \\

& & A: & SELECT Country FROM AIRPORTS WHERE AirportCode = 'AKO' \\
\hline
Unsupported Join & Schema & & Table albums, columns = ['*', 'albumId', 'singerId', 'albumName', 'yearReleased'] Table songs, columns = ['*', 'albumId', 'songId', 'songName'] Table singer, columns = [*,Singer\_ID,Name,Birth\_Year,Net\_Worth\_Millions,Citizenship] Table song, columns = [*,Song\_ID,Title,Singer\_ID,Sales,Highest\_Position] Foreign\_keys = [song.Singer\_ID = singer.Singer\_ID, songs.albumId = albums.albumId] \\
\cline{2-4}
& Conversation & U: & What is the name of the singer whose album is 'Thriller'? \\

& & A: & The albums and singer tables are not joined in the schema, so the original query cannot be executed. Could you please rephrase your question? \\

& & U: & What is the count of singers? \\

& & A: & SELECT count(*) FROM singer \\
\hline
Nonexistent Filter Value & Schema & & Table car\_makers, columns = [*,Id,Maker,FullName,Country] Table model\_list, columns = [*,ModelId,Maker,Model] Foreign\_keys = [model\_list.Maker = car\_makers.Id] \\
\cline{2-4}
& Relevant Values & & "car\_makers.FullName": ["General Motors", "Nissan Motors", "Kia Motors", "Ford Motor Company", "Triumph"] \\
\cline{2-4}
& Conversation & U: & How many car models were produced by the maker with full name American Motor Company? \\

& & A: & The full name "American Motor Company"does not exist in the car makers table. Could you please ask about another full name? \\

& & U: & What about "Triumph" instead? \\

& & A: & SELECT count(*) FROM CAR\_MAKERS AS T1 JOIN MODEL\_LIST AS T2 ON T1.Id  =  T2.Maker WHERE T1.FullName  = "Triumph"; \\
\hline
Nonexistent WHERE Column & Schema & & Table country, columns = [*, Code, Name, Region, SurfaceArea, IndepYear, Population, LifeExpectancy, GNP, GNPOld, LocalName, GovernmentForm, HeadOfState, Capital, Code2] \\
\cline{2-4}
& Conversation & U: & What is the total surface area of the continents Asia and Europe? \\

& & A: & There is no data on continents in the database. \\

& & U: & What is the total surface area of the countries in the Caribbean region? \\

& & A: & SELECT sum(SurfaceArea) FROM country WHERE Region = "Caribbean" \\
\bottomrule
\end{tabular}
\end{table*}

Table \ref{tab:ambi_example} and Table \ref{tab:unans_examples} show ambiguous and unanswerable examples from our dataset respectively.

\section{Human Annotation}
\label{appendix_human_annotation_procedure}

\subsection{Question Category Classification}

For question category classification, we sampled 20 questions from each category and ask SQL experts to classify whether the category is correct or not given the pair of modified question and database (that includes values from the tables retrieved for the filter criteria) as input (binary classification). We employ 2 SQL experts for the question category classification annotation task. Each of the annotators has at least a bachelor's degree in computer science. The annotators work as engineers/scientists in a private firm in the United States. The annotators performed their annotation task as a part of their service for which they were not specifically paid. To help with the annotations, we provide the definitions and a few examples of questions for each category.

\begin{figure*}
    \centering
    \includegraphics[width=0.8\textwidth]{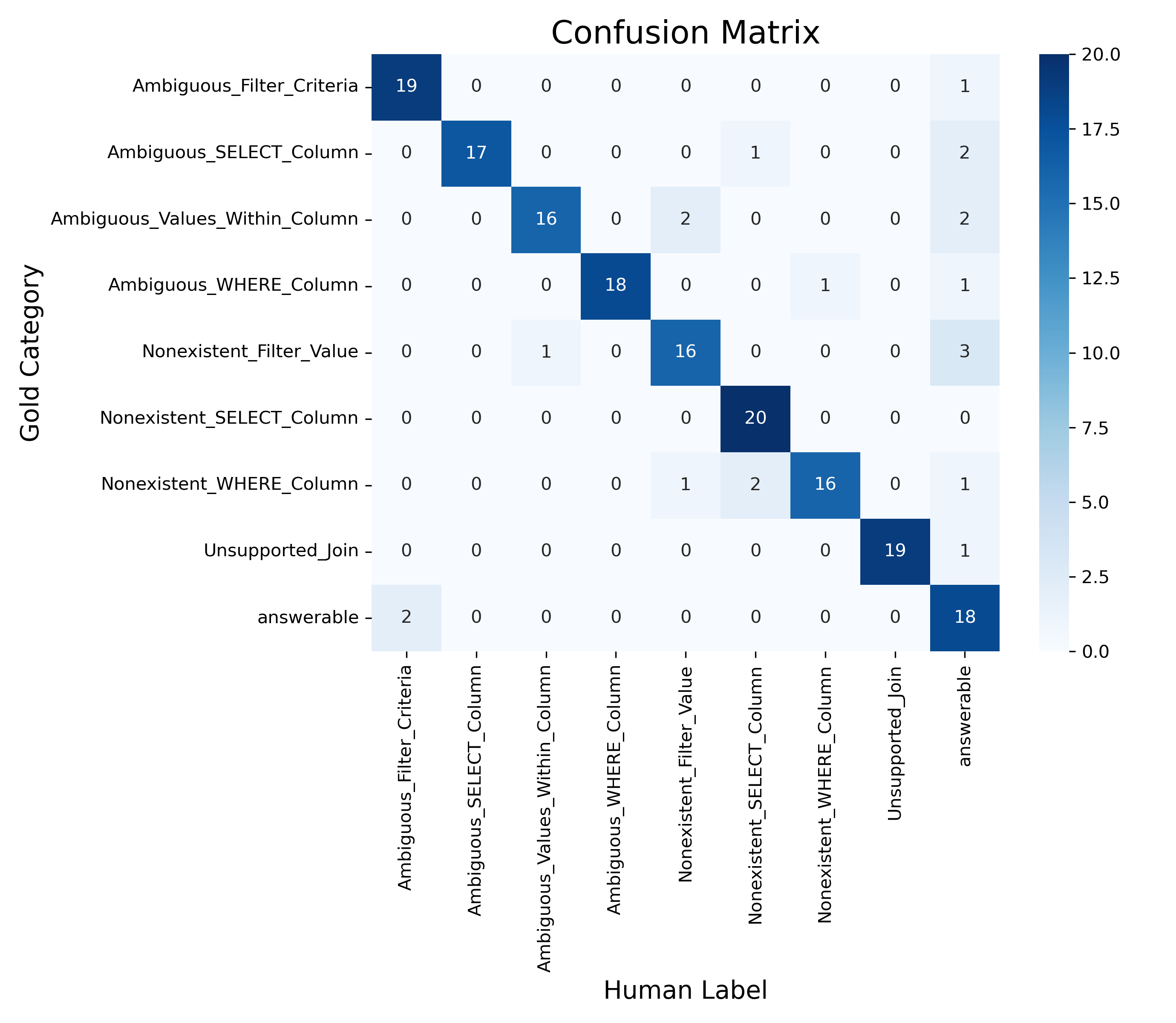}
    \caption{Figure showing the Confusion Matrix of Question Category Classification of the Human Annotation.}
    \label{fig:cm_human_evaluation}
\end{figure*}

Figure \ref{fig:cm_human_evaluation} shows the confusion matrix of the question category classification task of the human annotation. We see that in most cases the true label and the predicted labels are the same (diagonal entries in the matrix). Annotators classify ambiguous filter criteria, ambiguous where column, non-existent select column, unsupported join, and answerable categories with high accuracy. Nonexistent filter value is often classified as answerable mostly because annotators feel that the missing value is actually present in the schema and might not have been retrieved in the example provided. On the contrary, some answerable data is classified as ambiguous filter criteria, as the filter values might not have been retrieved properly causing the annotators to believe that the data belongs to ambiguous filter criteria. Nonexistent Where Column data is sometimes classified as Nonexistent Select Column as the annotators might believe that the column in the Select clause is missing for such examples. Ambiguous Values within Column is sometimes classified as Nonexistent Filter Value indicating that the ambiguous cell values are not retrieved and the annotators believe that the exact value is missing even though the value can be similar to multiple values in the database. Ambiguous Values within Column is also sometimes classified as Answerable because the annotators might mistakenly believe that the value required to answer the question is present in the database. Ambiguous Select Column is sometimes classified as answerable because the annotators might think that there exists another column apart from the column that is removed which can be used to answer the user question.  

\subsection{Conversation Quality Evaluation}
\label{ap:human_conv_annotation}
We sampled 90 conversations across ambiguous, unanswerable, and answerable categories from different databases for human annotation. Two SQL experts annotated each conversation on three criteria: factuality (correctness of SQL and natural language response), helpfulness (assistant's responses in understanding user intent), and naturalness (conversation flow) using a 1-5 Likert scale, where 1 denotes perfect/best quality and 5 denotes the worst quality.

Table~\ref{tab:summary_annotation_scores} shows the mean, standard deviation, and Krippendorff's Alpha for inter-annotator agreement. The high mean scores close to 1 (1.15-1.5) and substantial agreement (Alpha 0.68-0.82) indicate high-quality, natural conversations with factual and helpful responses. For Naturalness, we observe that annotators have a substantial agreement (Krippendorff's Alpha = 0.82), indicating that the conversations are generally perceived as natural and fluent. For Factuality, the annotators demonstrate moderate agreement (Krippendorff's Alpha = 0.68), suggesting that the conversations are consistently viewed as highly factual, which implies that the SQL queries in our dataset are of high quality. For Helpfulness, the annotators show good agreement (Krippendorff's Alpha = 0.76), indicating that the conversations are mostly helpful. Overall, 

Table~\ref{tab:human_annotation_results} presents category-wise annotation scores. Answerable data from Spider has a mean of 1 across criteria, confirming its high quality. Naturalness scores closer to 2 for most categories indicate mostly natural conversations, with 1 being perfectly natural. Categories like Ambiguous Filter Criteria, Nonexistent Filter Value, and Nonexistent Where Column have the most natural conversations (mean scores closer to 1), likely due to the close relation between user follow-up and assistant responses.

Factuality scores close to 1 across categories demonstrate accurate SQL generation and result descriptions through our reverse generation process, with 1 being perfectly factual. Helpfulness scores around 1.5 suggest mostly helpful assistant responses in understanding user intent, with 1 being perfectly helpful. Higher standard deviations for certain categories (e.g., Ambiguous Where Column, Ambiguous Values Within Column) indicate annotator disagreements due to varying relevance of ambiguous interpretations to user queries across examples.

Overall, the human annotation results validate the high quality, naturalness, factuality, and helpfulness of the generated conversational data.

\begin{table*}[!htb]
\centering
\small
\caption{Category-wise human annotation results for factuality, helpfulness, and naturalness. For every category, 1 denotes the best quality and 5 the worst.}
\label{tab:human_annotation_results}
\begin{tabular}{l|c c|c c|c c}
\hline
\textbf{Category} & \multicolumn{2}{c|}{\textbf{Naturalness}} & \multicolumn{2}{c|}{\textbf{Factuality}} & \multicolumn{2}{c}{\textbf{Helpfulness}} \\ 
 & \textbf{Mean} & \textbf{Std} & \textbf{Mean} & \textbf{Std} & \textbf{Mean} & \textbf{Std} \\ \hline
Ambiguous SELECT Column & 1.9 & 0.876 & 1.0 & 0.000 & 1.5 & 0.527 \\ \hline
Ambiguous WHERE Column & 1.9 & 0.994 & 1.1 & 0.316 & 1.6 & 1.075 \\ \hline
Ambiguous Values Within Column & 2.0 & 1.414 & 1.4 & 1.265 & 1.6 & 1.075 \\ \hline
Ambiguous Filter Criteria & 1.3 & 0.483 & 1.0 & 0.000 & 1.4 & 0.699 \\ \hline
Nonexistent Filter Value & 1.1 & 0.316 & 1.2 & 0.632 & 1.1 & 0.316 \\ \hline
Nonexistent WHERE Column & 1.35 & 0.412 & 1.2 & 0.483 & 1.35 & 0.474 \\ \hline
Unsupported Join & 1.7 & 0.919 & 1.1 & 0.316 & 1.5 & 0.972 \\ \hline
Nonexistent SELECT Column & 1.85 & 1.029 & 1.35 & 0.412 & 1.6 & 0.810 \\ \hline
Answerable & 1.0 & 0.000 & 1.0 & 0.000 & 1.0 & 0.000 \\ \hline
\end{tabular}
\end{table*}

\section{DIN-SQL Performance on Ambiguous and Unanswerable Queries}

As a probing task, we run DIN-SQL on a subset of our dataset containing ambiguous and unanswerable questions and analyze the results. The input to the DIN-SQL framework is an ambiguous/ unanswerable user query without the assistant response or the follow-up clarified user query. As expected, the model performs poorly on such data because the DIN-SQL framework is not designed to handle ambiguous and unanswerable user queries. During the schema linking, the model often hallucinates columns that do not exist in the database, potentially because the examples in the few-shot include only answerable questions.

\begin{table}[]
\scriptsize
\centering
\caption{Table showing DIN-SQL performance on ambiguous and unanswerable queries}
\label{tab:original-din-sql}
\begin{tabular}{p{2.3cm}|p{1.5cm}p{1cm}p{1cm}}
\hline
\textbf{Category} & \textbf{Hallucination (\%)} & \textbf{Incorrect SQL (\%)} & \textbf{Partially Correct (\%)} \\ \hline
Ambiguous SELECT Column & 3.8 & 3.8 & 92.4 \\ \hline
Ambiguous WHERE Column & 0 & 0 & 100 \\ \hline
Nonexistent SELECT Column & 45 & 55 & 0 \\ \hline
Nonexistent WHERE Column & 22 & 78 & 0 \\ \hline
Unsupported Join & 56 & 44 & 0 \\ \hline
\end{tabular}
\end{table}

Table \ref{tab:original-din-sql} shows the results of the DIN-SQL framework on ambiguous/ unanswerable user queries. Based on the results we make the following observations: 
\begin{itemize}
    \item For Ambiguous SELECT Column, we experiment with a total of 53 samples corresponding to different databases. We see that in two cases the framework hallucinates, i.e., it assumes that the removed column is actually present in the schema. In two other cases (Incorrect SQL) the framework predicts a completely different SQL that does not include the new Ambiguous SELECT Column names. In most cases 49/53, the framework predicts a partially correct SQL, i.e., it includes one of the Ambiguous SELECT Column names in its final prediction.
    \item For Ambiguous WHERE Column, we see that all the predicted SQLs contain one of the ambiguous columns and none of them are either hallucinating or missing the new columns.
    \item For Nonexistent SELECT Column, we see that 23 (45\%) cases lead to hallucination, i.e. the framework assumes that the removed column name is present in the schema and includes the column name in the final predicted SQL. 28 (55\%) cases predict an incorrect SQL (whose execution does not match the ground-truth execution results).
    \item For Nonexistent WHERE Column we see that only 7/31 (22\%) of cases lead to hallucination, i.e., the framework assumes that the removed column is still present in the schema and the rest 24/31 (78\%) are predicted as incorrect SQLs.
    \item In the case of Unsupported Join we see that 28/50 (56\%) of the SQLs are predicted with syntax errors/hallucinations where the framework assumes the presence of certain columns that do not exist in the schema to facilitate a JOIN operation to answer the question. 22/50 cases (44\%) have logical errors, in the predicted SQL i.e., they contain JOIN columns that do not have any foreign key relationship.
\end{itemize}

\section{Prompts}

\begin{figure*}
\centering
\caption{System prompt for generating replacement columns for Ambiguous SELECT Column data generation.}
\tiny
\begin{verbatim}
You will be provided with a database schema containing multiple tables and their columns.
The schema will be presented in a markdown format, along with 3 sample values for each column enclosed within <schema/> XML tags.
Additionally, you will be given a column of interest and its corresponding table within <column/> XML tags, a user question, and a corresponding SQL query.

Your task is to come up with two synonyms or phrases that have the same meaning as the original column of interest.
The goal is to remove the original column of interest and add the two new columns with similar values, making the user question ambiguous.
The synonyms should not be simple changes in case, pluralization, tense, etc. Instead, they should be alternative ways of expressing the same concept.

First, write your thought process within <scratch/> XML tags, following these steps:
1. Review the provided column of interest, its corresponding table, the user question, and the SQL query to understand the context.
    a. Identify the word/phrase mention that corresponds to the column of interest from the user question.
2. Brainstorm 5 potential synonyms or phrases for the column of interest. The synonyms or phrases shall have similar lexical overlap with the word/phrase mention in the question.
3. Evaluate each synonym/phrase and select the two best options that accurately capture the meaning of the original column, considering the following criteria:
    a. The synonyms and phrases should be similar to other columns within the schema in terms of wording, length, and style.
    b. A synonym or phrase is good if it is a valid and clear/obvious interpretation of the user question and results in a new SQL query that uses that interpretation.
    c. Write out why a synonym is a good explanation of the question and what the resulting new SQL will be if you think it is a good synonym. 
    Clearly explain how the synonym maps to the user question.
    d. If it is not a good synonym, explain why.

Then, write the two synonym columns within a Python list in the <result/> XML tags.
Each item within the list should be a dictionary containing the 'table' and 'column' keys, mapping to the respective table and column names.
The final two synonyms should have similar likelihoods of being the correct interpretations of what the original user question is referring to assuming the original column is interest is removed from the database.
The two synonyms or phrases shall have similar lexcial overlap with the mention in the question.
If you cannot find any suitable synonyms, output an empty Python list in the <result/> XML tags.
\end{verbatim}
\label{prompt:generating_replacement_select_columns}
\end{figure*}

\begin{figure*}
\centering
\caption{System prompt for generating user clarification response for Ambiguous SELECT Column data generation.}
\tiny
\begin{verbatim}
You will be presented with a database schema containing multiple tables and their columns.
The schema will be provided in a markdown format, along with 3 sample values for each column enclosed within <schema/> XML tags.
After the schema, you will see a conversation between a USER and a DB EXPERT within <conversation/> XML tags.
However, the follow-up question from the user before the final SQL query will be missing.

Your task is to generate a natural, logical, and concise follow-up USER question based on the final SQL query provided by the DB EXPERT.

Write your thinking process within <scratch/> XML tags, following these steps:
1. Carefully review the database schema to understand the structure and relationships between the tables.
2. Analyze the final SQL query to comprehend the information it retrieves and the operations it performs.
3. Formulate a follow-up question that accurately reflects the intent and results of the final SQL query, without any unnecessary or redundant words.
4. Ensure that the follow-up question is natural and free from unnatural phrases, such as phrases with underlines or unnatural casing.

In the end, write the follow-up question within the <result/> XML tags without any additional explanations.
\end{verbatim}
\label{prompt:generating_user_clarification_response}
\end{figure*}

\begin{figure*}
\centering
\caption{System prompt for refining the user's follow-up/clarification response/question.}
\tiny
\begin{verbatim}
You will be presented with a database schema containing multiple tables and their columns.
The schema will be provided in a markdown format, along with 3 sample values for each column enclosed within <schema/> XML tags.
After the schema, you will see a conversation between a USER and a DB EXPERT within <conversation/> XML tags.
The initial user question is often ambiguous or unanswerable, and the DB EXPERT explains why. 
The user then asks a follow-up question that is answerable but verbose.

Your task is to rephrase the user's verbose follow-up question. 
The rephrased question should convey the requested clarification (such as which column or cell value to use) in a concise, conversational, and natural way based on the context provided by the initial ambiguous question and the DB expert's explanation. 
It is important not to omit any information where the DB expert has requested clarification.

The rephrased follow-up question should be clear, logical, and easy to understand, while avoiding unnecessary repetition of information from the initial conversation and technical jargon or complex words. 
Do not include unnecessary filler words like "hey" or "hello".

First, think step by step in <scratch/> XML tags.
Then, write the rephrased concise follow-up question within the <result/> XML tags without any extra explanation.
\end{verbatim}
\label{prompt:refine_conversation_prompt}
\end{figure*}

\begin{figure*}
\centering
\caption{System prompt for adding execution results explanation based on the SQL execution results.}
\tiny
\begin{verbatim}
You will be provided with a database schema containing multiple tables and their columns in a markdown format, along with 3 example values for each column enclosed within <schema/> XML tags.

After the schema, there will be a conversation between a USER and a DB EXPERT within <conversation/> XML tags.
The conversation will include the USER's query and the DB EXPERT's SQL query to retrieve the requested information. 
At the end, the execution results of the DB EXPERT's final SQL query will be presented within <execution_results/> XML tags.

Your task is to analyze the execution results and provide a short answer to the USER's last question based on those results. 
The answer should be concise, self-contained, and no more than two sentences. 
Do not comment on the correctness of the query itself. Note that the short answer shall be self-explanatory becaues USER only sees the answer but not the execution results.

Your answer should be clear, logical, and easy for the USER to understand. Avoid using technical jargon or overly complex explanations. 
The goal is to provide a helpful and informative answer to the USER's question.

Note that if the execution results contain more than 30 rows, only the first 30 rows will be shown, and the remaining rows will be skipped.

First, outline your thought process step by step within <scratch/> XML tags. Then, write the rephrased response within the <result/> XML tags without any additional explanation.
\end{verbatim}
\label{prompt:generate_execution_results_explanation}
\end{figure*}

\begin{figure*}
\centering
\caption{Binary classification Prompt for data filtering. "\{category\_with\_explanation\}" will be replaced with the name and definition of the corresponding question category in Figure \ref{prompt:category_to_explanation_mapping}. Few-shots examples are presented as conversation between the user and assistant in the format of message API of litellm.}
\tiny
\begin{verbatim}
You are a Database Expert (DB EXPERT) system that classifies user questions into one of the following two categories based on the given database schema:

{category_with_explanation}

- answerable: the database contains data needed to answer the question and the question has one and only one valid interpreation.

You will receive:
1. A database schema in markdown format with relevant column values enclosed in <schema/> tags.
2. The user's question enclosed in <question/> tags.

Note that the "answerable" output shall only be provided if and only if:
- the question posed is unambiguous, precise, and leaving no room for multiple interpretations or confusion. 
- the database contains the complete set of information required to formulate a comprehensive and accurate response to the query. 
If either of these conditions is not met, meaning the question lacks clarity or our data is insufficient, we shall refrain from classifying the query as "answerable."

Provide your step-by-step thoughts within </scratch> tags.
Then, provide your final classification within <result/> tags as one of the categories above.
Note that you result shall only be one of the categories specified at the beginning & Do not include any extra explanation in the result. 
\end{verbatim}
\label{prompt:binary_classification}
\end{figure*}

\begin{figure*}
\centering
\caption{Nine-way classification System Prompt. "\{category\_with\_explanation\}" will be replaced with the name and definition of four ambiguous and four unanswerable categories in Figure \ref{prompt:category_to_explanation_mapping}. Few-shots examples are presented as conversation between the user and assistant in the format of message API of litellm.}
\tiny
\begin{verbatim}
You are a Database Expert (DB EXPERT) system that classifies user questions into one of the following 9 categories based on the given database schema:

{category_with_explanation}

- answerable: the database contains data needed to answer the question and the question has one and only one valid interpreation.

You will receive:
1. A database schema in markdown format with relevant column values enclosed in <schema/> tags.
2. The user's question enclosed in <question/> tags.

Your output should follow this format:
<scratch> YOUR-STEP-BY-STEP-THOUGHTS </scratch>
<result> ONE-OF-THE-9-QUESTION-CATEGORIES </result>

Note that the "answerable" output shall only be provided if and only if:
- the question posed is unambiguous, precise, and leaving no room for multiple interpretations or confusion. 
- the database contains the complete set of information required to formulate a comprehensive and accurate response to the query. 
If either of these conditions is not met, meaning the question lacks clarity or our data is insufficient, we shall refrain from classifying the query as "answerable."

Provide your step-by-step thoughts within </scratch> tags.
Then, provide your final classification within <result/> tags as one of the categories above. Do not include any extra explanation in the result.

\end{verbatim}
\label{prompt:nine_way_classification}
\end{figure*}

\begin{figure*}
\centering
\caption{Ambiguous and Unanswerable category and their definition. These definitions are used in the binary and 9-way classification.}
\tiny
\begin{verbatim}
category_to_explanation_mapping = {
"Ambiguous_SELECT_Column": 
  "Multiple columns match the requested output information, leading to multiple valid SQLs that differ in the columns used in the SELECT clause.",
"Ambiguous_WHERE_Column": 
  "The filter condition matches multiple columns in a table, leading to multiple valid SQLs that differ in the specific filter column in the WHERE clause.",
"Ambiguous_Values_Within_Column": 
  "Multiple similar values in a column match the mentioned value, leading to multiple valid SQLs that differ in the specific filtering cell value from one column in the WHERE clause.",
"Ambiguous_Filter_Criteria": 
  "The question contains a filter condition or criteria that is ambiguous, vague, relative/descriptive, or open to multiple interpretations."
  "This ambiguity in the filter criteria makes it difficult to formulate a precise SQL query without additional clarification.",
"Nonexistent_SELECT_Column": 
  "At least one of the requested output information is not present in the schema, so no SQL can be constructed with the appropriate column in the SELECT clause.",
"Nonexistent_WHERE_Column": 
  "At least one filter condition column is not present in the schema, so no SQL can be constructed with the appropriate column in the WHERE clause",
"Unsupported_Join": 
  "The required join between tables is not supported due to a lack of common columns, preventing the construction of a valid SQL query",
"Nonexistent_Filter_Value": 
  "The mentioned filtering value is not present in the schema, so no SQL can be constructed with the appropriate value in the WHERE clause.",
}

\end{verbatim}
\label{prompt:category_to_explanation_mapping}
\end{figure*}

\begin{figure*}
\centering
\caption{Example user message and assistant message that are used as fewshots in the question category classification.}
\tiny
\begin{verbatim}
User:

<schema>

## shop

| Column Name | Data Type | Description |
| --- | --- | --- |
| Location | str | Example values: Dinas, Pentre, Porth, Cymmer |
| Open_Date | str | Example values: 11 April, 10 December, 1 January, 10 June |
| Open_Year | str | Example values: 2014, 2006, 2009, 2011 |
| Shop_ID | int | Example values: 1, 2, 3, 4 |
| Shop_Name | str | Example values: Dinas Device, Pentre store, Dinas store, Best Buy |

## stock

| Column Name | Data Type | Description |
| --- | --- | --- |
| Device_ID | int | Example values: 4, 6 |
| Quantity | int | Example values: 100, 110, 134, 56 |
| Shop_ID | int | Example values: 1, 2, 3, 4 |

## device

| Column Name | Data Type | Description |
| --- | --- | --- |
| Applications | str | Example values: 5.0.0.419, 5.0.0.328, 4.7.0.151, 5.0.0.451 |
| Device | str | Example values: Huawei, Apple, BlackBerry Storm 9530, Xiaomi |
| Device_ID | int | Example values: 1, 2, 3, 4 |
| Package_Version | str | Example values: 5.0.0.808, 5.0.0.328, 5.0.0.419, 4.7.0.208 |
| Software_Platform | str | Example values: Android, iOS |



</schema>

<question>
What are the carriers of devices that are not in stock anywhere?
</question>

-------------------------------------------------------------------------------
Assistant:

<scratch>
To determine if the question is answerable or not, I need to check if the requested information (carriers of devices) is present in the given schema.

The schema contains three tables:
1. shop: Contains information about shop locations, opening dates, and IDs.
2. stock: Contains information about device IDs, quantities, and shop IDs.
3. device: Contains information about device details like applications, device name, device ID, package version, and software platform.

The question is asking for the "carriers" of devices that are not in stock anywhere. However, there is no column in the schema that contains information about carriers.

The closest information we have is the "Device" column in the "device" table, which contains the device names like "Huawei", "Apple", "BlackBerry Storm 9530", and "Xiaomi".
But these are not carriers, they are device manufacturers/brands.

Since the requested information (carriers) is not present in the schema, this question falls under the "Nonexistent SELECT Column" category.
There are no SQL query can be constructed with the appropriate column in the SELECT clause to retrieve the requested information.
</scratch>

<result>
Nonexistent_SELECT_Column
</result>

\end{verbatim}
\label{prompt:example_user_assistant_message}
\end{figure*}

\section{Dataset Construction for each ambiguous/unanswerable category}
\label{appendix-sec:dataset-construction}

In this section, we describe the detailed procedure for constructing data for each ambiguous/unanswerable category as described in the Ambiguous SELECT Column. 

\subsection{Ambiguous WHERE Column} 

In stage 1, we collect the column names appearing in the Where clause of the SQLs of all questions in the Spider dataset. Like in the Ambiguous SELECT Column case, we then provide those columns as inputs to an LLM and prompt it to generate column names that are semantically similar but not equivalent. We then construct the value of ambiguous problematic data by removing the original column from the schema and adding the newly generated Ambiguous SELECT Columns. In stage 2, we construct the assistant's helpful response using a template similar to the Ambiguous SELECT Column case. We generate the assistant's clarified SQL by replacing the column in the Where clause of the original SQL with one of the Ambiguous SELECT Columns. We then provide this data in the form of a conversation as input to the LLM and prompt it to come up with the user clarification response. Like in the Ambiguous SELECT Column case, after generating the user clarification we filter the data based on some rules. We discard those samples that contain the new column or miss the removed column in the clarification. We finally refine the conversation, execute the clarification SQL to get the results, and generate the natural language explanation of the results. 

\subsection{Ambiguous Values Within Column}

In stage 1, we extract the values appearing in the Where clause of the SQLs of all questions in the Spider dataset. We prompt the LLM to generate values that are similar but not equivalent to each other. We then construct the Ambiguous Values Within Column ambiguous data by constructing a new schema from the original schema by removing the original value and adding the newly generated ambiguous values. For example, for the value ``chemistry'' the LLM generates two ambiguous values, ``organic chemistry'' and ``physical chemistry''. In stage 2, we construct the assistant's helpful response by using a template that points out the presence of two Ambiguous SELECT Columns. We generate the assistant's clarified SQL by replacing the original value with one of the ambiguous values. We then provide this data in the form of a conversation as input to the LLM and prompt it to come up with the user clarification response. We then discard those data where the user clarification does not mention the newly generated values. Finally, we refine the conversation, execute the clarification SQL get the results, and generate the natural language explanation of the results.

\subsection{Ambiguous Filter Criteria}

To construct the ambiguous filter criteria data, we utilized the SPIDER dataset. Instead of modifying the databases, we prompted a Large Language Model (LLM) to modify the user questions to introduce ambiguity. Specifically, we employed the following techniques:
1. Replacing specific filter values with relative terms like "little/large," "young/old," "slow/fast," etc.
2. Using descriptive terms instead of explicitly stating the original filter value.
The modified questions resembled those in the BIRD dataset that require additional "evidence" or definitions to convert text to SQL. After modifying the questions, we prompted the LLM with different instructions to generate a response from the database assistant's perspective, explaining why the question was ambiguous.
Finally, we used the original (unmodified) user question as the clarified follow-up question, and the corresponding SQL as the gold SQL after the user's clarification.

\subsection{Nonexistent SELECT Column}

In stage 1, we extract the columns appearing in the Select clause of the SQLs of all questions in the Spider dataset and construct new schemas by removing the columns required for answering the respective questions. In stage 2, we construct the assistant's helpful response using a template that states that the column required for answering the question is missing from the schema. We construct the final SQL by replacing the missing column from the schema (in the Select clause) with a column that exists in the schema. We convert this data into conversational data and prompt the model to generate the user clarification response. In stage 3, we refine the conversation, execute the clarification SQL get the results, and generate the natural language explanation of the results. 

\subsection{Nonexistent Filter Value}

In stage 1, we extract the values appearing in the Where clause of the SQLs of all questions in the Spider dataset. For constructing the problematic data we construct a new schema by removing the values required for answering the question from the schema. In stage 2, we construct the assistant's helpful response using a template that mentions that the value mentioned in the question is not present in the schema. We construct the clarification SQL by replacing the removed value with another value present in the schema. We then convert this data into conversational data and prompt the LLM to generate the user clarification response. In stage 3, we refine the conversation, execute the clarification SQL get the results, and generate the natural language explanation of the results. 

\subsection{Unsupported Join}

In stage 1, to construct the problematic data, we consider the unique schemas of the Spider dataset and prompt the LLM to generate a new schema with at least two new tables and corresponding columns such that the new tables have a foreign key relationship with themselves but not with any other column in the schema. For example, for a schema containing student information like student grade, teacher details, etc. the LLM produces two new tables of library and books that have a foreign key relationship with each other but not with any other table in the original schema. In stage 2, we construct the assistant's helpful response using a template stating that the question requires joining tables of the schema that have no relationship with each other. We construct a clarification SQL by using SQL from the Spider dataset corresponding to the original schema. We then convert this data into conversational data and prompt the LLM to generate the user clarification response. In stage 3, we refine the conversation, execute the clarification SQL get the results, and generate the natural language explanation of the results. 

\subsection{Nonexistent WHERE Column}

In stage 1, we extract the columns present in the Where clause of the SQLs in the Spider dataset and construct new schemas by removing the columns required for answering the respective questions. In stage 2, we construct the assistant's helpful response using a template that mentions that the information required for answering the question is not present in the schema. We construct the clarification SQL by finding a SQL from the Spider dataset whose Select columns match the problematic question and whose Where columns are present in the schema. We convert this data into a conversational format and prompt the LLM to generate the user clarification response. In stage 3, we refine the conversation, execute the clarification SQL get the results, and generate the natural language explanation of the results.

\section{Experimental Settings}

We use Anthropic AI's Claude 3 Sonnet via Amazon Bedrock \footnote{\url{https://www.anthropic.com/}, \url{https://aws.amazon.com/bedrock/}} for all our data generation. For the zero-shot and the few-shot prompts designed for evaluating the dataset, we use Claude 3 Sonnet, Haiku, Llama-3.1 70B, and LLama-3-1-8B with a greedy decoding strategy, i.e., we set the \emph{top-p} value to 1.0 and temperature to 0.0. We implement the DIN-SQL model using Claude 3.5 Sonnet, Claude 3 Sonnet, Llama3-1-70B, Llama3-1-8B, and Mixtral-Large-2 by tailoring the original GPT-4 based prompts and using the same set of hyperparameters as that used by GPT-4. 
Future work can focus on evaluating our dataset with the DIN-SQL model implemented using GPT-4.

\section{Dataset Access and Distribution}
We will make the code and prompt used to used for generating and benchmarking the data open-source under the MIT License\footnote{\url{https://opensource.org/license/mit/}} for the community to access and contribute. We use the open-source Spider\footnote{We only used ``train\_spider.json'' and ``dev.json'' from Spider in our dataset construction and related experiments. We did NOT use any data from ``train\_others.json'' or ``yelp''.} dataset for creating \emph{PRACTIQ}. The Spider dataset is governed by CC BY-SA 4.0 license which allows us to freely use the data for modification. To the best of our knowledge, we make sure that the dataset does not contain the private information of any individual or entity.

\end{document}